\documentclass[runningheads]{llncs}
\usepackage[T1]{fontenc}
\usepackage{graphicx}
\usepackage{url}
\begin{document}
\title{Virtual Patients, Real Gains: Simulated CT from Digital Twins for Multi-Task Lung Nodule Analysis}
\titlerunning{Simulated CT from Digital Twins for Lung Nodules}
%
\author{Fakrul Islam Tushar\inst{1,2}\and
Lavsen Dahal\inst{2} \and
Paul Segars\inst{2} \and
Joseph Y. Lo\inst{2}}
\authorrunning{F. I. Tushar et al.}
\institute{Department of Radiology and Imaging Sciences, University of Arizona
\and
Center for Virtual Imaging Trials, Department of Radiology, Duke University Medical Center}
  
\maketitle              
\begin{abstract}
AI-based lung cancer screening is constrained by scarce, annotated CT data, particularly for rare nodule presentations. We investigate whether physics-based, anatomy-informed simulated CT can improve AI performance across three lung-nodule tasks: detection, segmentation, and malignancy classification. Using the Virtual Lung Screening Trial framework, we generated 174 digital human twins (XCAT3), embedded 512 procedurally controlled nodules (X-Lesions, 4--30 mm), and simulated CT (DukeSim) under two scanner configurations, yielding 1,044 annotated scans. Combined with clinical data, these trained models for detection (MONAI), segmentation (VISTA3D, nnU-Net), and classification (Med3D), evaluated on external test sets. Detection sensitivity at 1 FP/scan rose from 0.37 to 0.56 ($p<0.001$); segmentation improved modestly (Dice 0.61$\rightarrow$0.64, 0.66$\rightarrow$0.69); classification AUC rose from 0.78 to 0.87 ($p<0.001$). Physics-based virtual imaging trials can help address data scarcity in medical AI.


\end{abstract}
\section{Introduction}

\begin{figure}[!t]
    \centering
    \begin{minipage}[t]{0.48\textwidth}
        \centering
        \includegraphics[width=\textwidth]{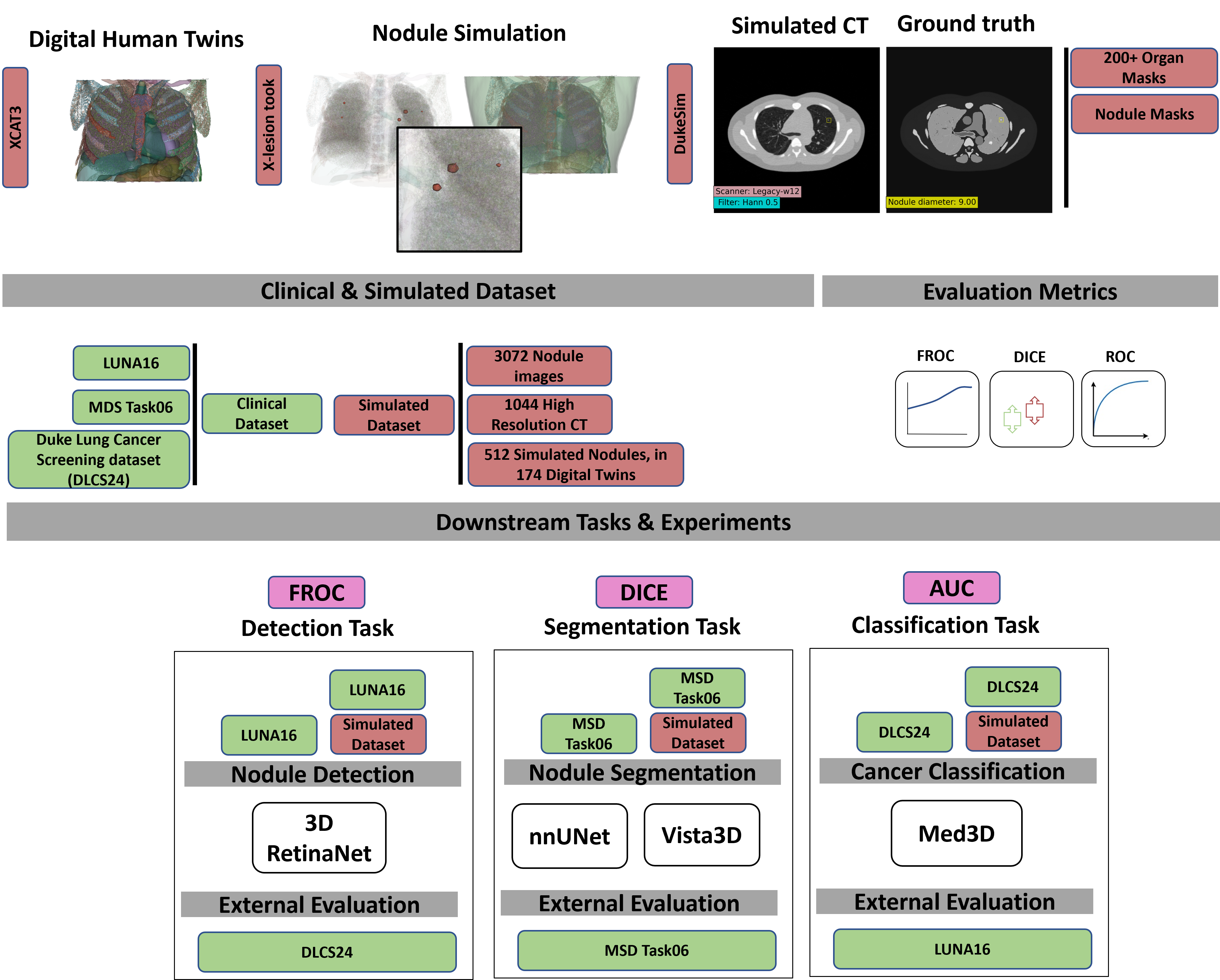}
        \caption{Overview of the workflow, integrating XCAT3, X-Lesion, and DukeSim for digital twin and nodule simulation. Clinical and simulated datasets are used for nodule detection, segmentation, and classification with external evaluations using FROC, Dice, and AUC.}
        \label{study_overview}
    \end{minipage}\hfill
    \begin{minipage}[t]{0.48\textwidth}
        \centering
        \includegraphics[width=\textwidth]{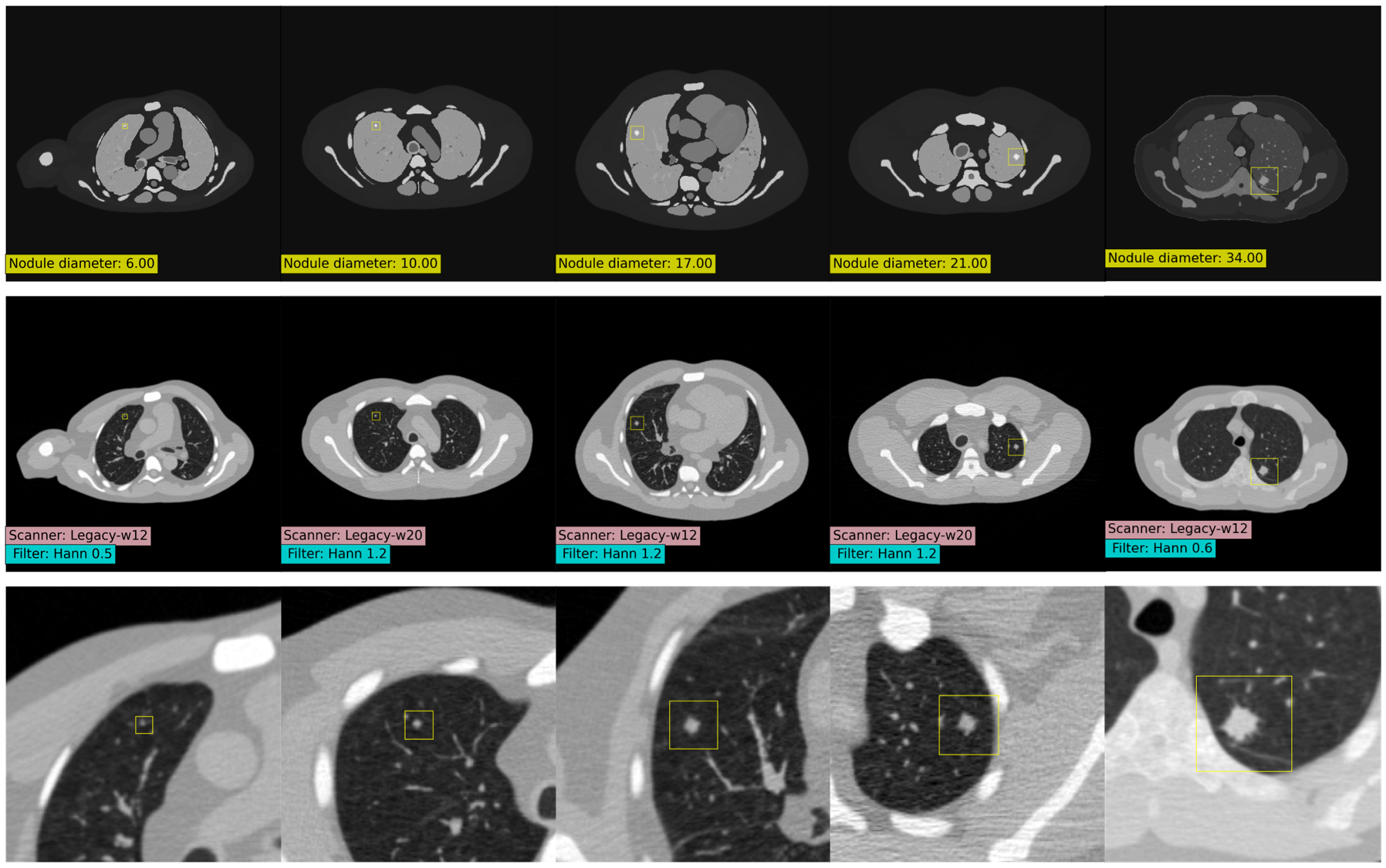}
        \caption{Simulated lung nodules with varying sizes and imaging conditions. Top: digital human twins model slices with embedded nodules. Middle: simulated CT with different scanner settings. Bottom: zoomed-in nodule views.}
        \label{SYNLUNGS_EXAMPLE}
    \end{minipage}
\end{figure}

Deep learning now underpins computer-aided detection and diagnosis for lung cancer screening, yet its performance is bounded by data scarcity, especially for rare presentations~\cite{he2025vista3d,guo2025maisi}. Lung nodules vary widely in size, shape, and type, but real-world datasets rarely capture this diversity and often lack segmentation masks and diagnostic labels~\cite{national2011reduced,setio2017validation,wang2025duke,tushar2024ai,antonelli2022medical}. Expert annotation is costly and slow, and privacy constraints further limit data sharing.

Diffusion-based generative models have emerged as one response, showing promise for medical image synthesis~\cite{guo2025maisi,tushar2025nodmaisi}. Although promising, they often lack anatomical and pathological constraints, failing to capture the full diversity of lung nodules or rare cases of disease~\cite{tushar2025nodmaisi}. Furthermore, most generative synthesis studies do not explicitly model clinically relevant imaging scenarios, such as scanner-specific acquisition physics or reconstruction protocol variations that significantly impact image characteristics. Without these constraints, generative models risk propagating the biases in their training data while failing to deliver the controlled diversity that AI development demands.

An alternative paradigm leverages computational phantoms: digital representations of human anatomy that serve as controlled, reproducible tools in imaging research~\cite{abadi2020virtual}. Physical anthropomorphic phantoms have long supported system evaluation, image-quality assessment, and protocol optimization~\cite{glick2018advances,akhavanallaf2022update,black2021design,followill2007design}. Such models have progressed from single-organ simulations, as in breast imaging, to whole-body twins that reproduce complex anatomy with high fidelity~\cite{abadi2020virtual,abadi2018dukesim,dahal2025xcat,badano2018evaluation,tushar2025virtual}. Integrated with physics-based imaging simulation, these digital human twins support comparative evaluation of imaging technologies, disease quantification, and in silico clinical trials~\cite{abadi2020virtual,badano2018evaluation,tushar2025virtual}; more recently, virtual imaging trials (VITs) have shown promise for evaluating AI-based diagnostic systems by providing well-annotated ground truth otherwise difficult or impossible to obtain clinically~\cite{tushar2025virtual,abadi2017airways,sauer2023surface,tushar2026utilityvirtualimagingtrials}.

Building on these advances, we ask whether physics-based, anatomy-informed simulated CT can strengthen AI training for lung cancer diagnosis. We draw on a validated simulated dataset from the Virtual Lung Screening Trial (VLST)~\cite{tushar2025virtual}, which integrates digital human twins (XCAT3 phantoms), procedurally generated synthetic lung nodules (X-Lesions)~\cite{sauer2019modeling,mccabe2024synthesizing,tushar2025virtual}, and physics-based CT simulation (DukeSim)~\cite{abadi2018dukesim}. Enriching clinical training data with this simulated set, we assess performance and generalization across three fundamental tasks: nodule detection, segmentation, and malignancy classification; comparing clinical-only against clinical + simulated configurations, with all evaluation on clinical test sets spanning diverse imaging protocols and patient populations.

To our knowledge, this is the first study to evaluate physics-based virtual imaging data for multi-task AI development in lung cancer screening, using internal and external validation to establish whether virtual imaging can address data-scarcity limitations in AI training.

\section{Methods}

Fig.~\ref{study_overview} summarizes the end-to-end workflow, including digital twin generation, nodule insertion, CT simulation, and downstream evaluations. The following sections provide detailed descriptions of digital human modeling, nodule simulation, imaging simulation, and the final dataset composition.

\subsection{Clinical Datasets}
Clinical datasets were strategically selected to provide comprehensive evaluation across detection, segmentation, and classification tasks while representing diverse imaging protocols, annotation types, and clinical populations.

The \textbf{Duke Lung Cancer Screening (DLCS24)}~\cite{wang2025duke} dataset comprises 1,613 patients with 2,487 annotated lung nodules from the Duke University Health System. Each nodule is marked with a 3D bounding box (center coordinates x, y, z; width, height, and depth) linked to clinical and pathological outcomes. Annotations were performed by a medical student supervised by cardiothoracic imaging radiologists, following Lung-RADS v2022 criteria to identify nodules measuring at least 4 mm or located in central or segmental airways.
The cohort includes patients with mean age 66 years (range: 50--89), 50.3\% male, and racial distribution of 74.1\% White, 22.7\% Black/African American, and 3.2\% Other/Unknown. Among the annotated nodules, 2,223 (89.4\%) were benign with 264 (10.6\%) confirmed malignant through histopathology and clinical follow-up. For our study, we used the publicly available subset with predefined train/validation/test splits~\cite{tushar2024ai}. The DLCS24 dataset provides both 3D bounding box annotations for detection and clinically adjudicated malignancy labels for classification, making it suitable for multi-task evaluation. The dataset is publicly available via Zenodo \url{https://zenodo.org/records/13799069}.

The \textbf{Lung Nodule Analysis 2016 (LUNA16)}~\cite{setio2017validation} is a widely used benchmark dataset comprising 888 CT scans derived from the Lung Image Database Consortium and Image Database Resource Initiative (LIDC-IDRI)~\cite{armato2011lung}, refined to 601 scans with 1,186 annotated nodules for the challenge. Each nodule annotation includes 3D bounding box coordinates (center x, y, z) and diameter measurements. The dataset employs a predefined 10-fold cross-validation protocol for standardized evaluation.
For nodule detection, we followed the official LUNA16 challenge protocol. For malignancy classification, LUNA16 lacks histopathologically confirmed outcomes; therefore, we adopted the Radiologist Suspicion Label (RSL) proxy reference standard established in prior work~\cite{pai2024foundation,tushar2024ai}, which retains 677 nodules where at least one radiologist indicated malignancy suspicion based on LIDC-IDRI's five-point ordinal malignancy scoring system. The dataset is available at \url{https://luna16.grand-challenge.org/Data/}.

The \textbf{Medical Segmentation Decathlon Task06 (MSD Task06)}~\cite{antonelli2022medical} lung dataset, part of the Medical Segmentation Decathlon challenge, consists of 96 thin-section CT scans from patients with non-small cell lung cancer (NSCLC). The dataset is officially partitioned into 64 training cases and 32 test cases, though the publicly available training set contains 63 cases. MSD Task06 includes voxel-wise 3D segmentation masks for lung tumors, making it particularly valuable for training and evaluating segmentation algorithms. The dataset is available through the MSD repository (\url{http://medicaldecathlon.com/}).

\subsection{Simulated Dataset: Digital Twins, Virtual Population, and Imaging}

We adopted the validated virtual population from the VLST, an established in silico imaging study, as the foundation for our simulated CT dataset~\cite{tushar2025virtual}. This cohort provides 174 anatomically diverse digital human twins with 512 embedded synthetic lung nodules. The VLST dataset simulated CT images under multiple scanner configurations, providing 1,044 scans with complete 3D ground truth annotations for AI model training.

\textbf{XCAT3 phantoms}~\cite{dahal2025xcat} were used to develop digital human twins, curating anatomical models from clinical chest CT scans. Instead of direct image synthesis, this structured pipeline integrates AI-driven segmentation and procedural texture modeling to generate anatomy-informed representations. The process began with nnU-Net segmentation using DukeSeg to delineate over 200 anatomical structures. A multi-step quality control process ensured anatomical fidelity, including statistical validation, anomaly detection, and physician review. Following segmentation, procedural modeling generated tissue textures such as lung vasculature, airways, and trabecular bone structures~\cite{abadi2017airways,sauer2023surface}. Each tissue type was mapped to its material properties to facilitate the next step for physics-based simulations. The axial slices of the digital human twins are shown in Fig.~\ref{SYNLUNGS_EXAMPLE}.

\textbf{X-Lesions tool}~\cite{sauer2019modeling,mccabe2024synthesizing} was utilized to generate and embed simulated lung nodules into digital human twins. Nodules were created at 0.1 mm voxel resolution, with controlled size, shape, and density. Internal heterogeneity of texture was introduced by the 3D clustered lumpy background (CLB) model~\cite{mccabe2024synthesizing}. Lesion sizes were drawn from a distribution based on National Lung Cancer Screening (NLST)~\cite{national2011reduced} nodule data. Fig.~\ref{SYNLUNGS_EXAMPLE} shows axial slices of digital twins with embedded nodules. After nodule insertion, CT imaging was simulated with \textbf{DukeSim}~\cite{abadi2018dukesim}, a validated tool using ray tracing for primary signals and Monte Carlo simulation for scatter and dose. Projections were reconstructed with a vendor-neutral MCR toolkit replicating two scanners: a W12 configuration (12 mm collimation, narrower 1.5 mm detectors) and a W20 configuration (20 mm collimation, wider 2.19 mm detectors). Both acquired 1,000 projections per scan across three reconstruction filters (Hann 0.5, 0.6, 1.2), spanning a representative range of acquisition conditions. Fig.~\ref{SYNLUNGS_EXAMPLE} presents examples of simulated CT images.

\textbf{The simulated dataset} consists of 174 digital human twins with 512 simulated lung nodules, generating \textbf{1,044 CT scans and 3,072 nodule images} across the two scanners. Each scanner contributed 522 CT scans and 1,536 nodule images, ensuring coverage of diverse imaging conditions. The dataset includes 95 males (54.6\%), ages 59 $\pm$ 15 years, with a mean BMI of 26 $\pm$ 6. The cohort is 73.6\% White, 20.7\% Black, and 5.8\% Other/Unknown, with 98.3\% identifying as non-Hispanic. Unlike LUNA16~\cite{setio2017validation} and DLCS24~\cite{wang2025duke,tushar2024ai}, which provide 3D bounding boxes, only MSD Task06~\cite{antonelli2022medical} and simulated dataset include 3D segmentation masks. While NLST~\cite{national2011reduced} lacks segmentation, it provides malignancy labels. Simulated nodules incorporate statistically derived malignancy labels, adapted from prior studies~\cite{tushar2024beyond}. Among the 174 digital twins, 124 were used for AI model training, while 50 were reserved as a test set.

\begin{table}[!t]
\centering
\caption{Dataset Characteristics and Usage Across Detection, Segmentation, and Classification Tasks}
\label{tab:dataset_characteristics}

\scriptsize
\setlength{\tabcolsep}{2pt}
\renewcommand{\arraystretch}{1.1}

\begin{tabular}{|p{2.1cm}|p{2.4cm}|p{2.6cm}|p{1.9cm}|p{2.4cm}|}
\hline
\textbf{Task} &
\multicolumn{3}{c|}{\textbf{Training Data}} &
\textbf{Test Data} \\
\cline{2-4}
& \textbf{Clinical Dataset} & \textbf{Simulated Dataset} & \textbf{Combined} & \textbf{Clinical Test} \\
\hline

\textbf{Detection} &
LUNA16~\cite{setio2017validation}: 601 CTs; 1,186 nodules. &
744 CTs; 2,250 nodules. &
1,345 CTs; 3,436 nodules &
DLCS24~\cite{wang2025duke,tushar2024ai}: 198 CTs; 294 nodules (test split) \\
\hline

\textbf{Segmentation} &
MSD Task06~\cite{antonelli2022medical}: 50 CTs, 51 nodules. &
744 CTs; 2,262 nodules (377 $\times$ 6 imaging configurations) &
794 CTs; 2,312 nodules. &
MSD Task06: 13 CTs, 13 nodules \\
\hline

\textbf{Classification} &
DLCS24~\cite{wang2025duke,tushar2024ai}: Train 1,618 nodules (1,452 benign, 166 cancer); validation 575 nodules. &
366 nodules (statistically labeled: 327 benign, 39 cancer) &
Train 1,984 nodules; validation 575 nodules &
LUNA16~\cite{pai2024foundation}: 677 nodules with RSL proxy labels \\
\hline
\end{tabular}

\vspace{0.6ex}
\parbox{0.98\linewidth}{\scriptsize
\textit{Note:} The simulated dataset was generated from the VLST framework: 174 digital twins with 512 embedded nodules.
For segmentation, 124 twins (377 nodules) were used for training, with 50 twins (135 nodules) reserved for internal testing.
Each twin was imaged with 2 scanner configurations $\times$ 3 reconstruction filters = 6 imaging variants.
}
\end{table}

\section{Experiments and Results}
We evaluated simulated-data enrichment across three tasks: detection, segmentation, and malignancy classification. For each, we trained a clinical-only baseline and a clinical + simulated model with identical settings, evaluating both on external clinical test sets. Table~\ref{tab:dataset_characteristics} summarizes dataset composition, training configurations, and test sets across all tasks.

\begin{figure}[!t]
    \centering
    \begin{minipage}[t]{0.58\textwidth}
        \centering
        \includegraphics[width=\textwidth]{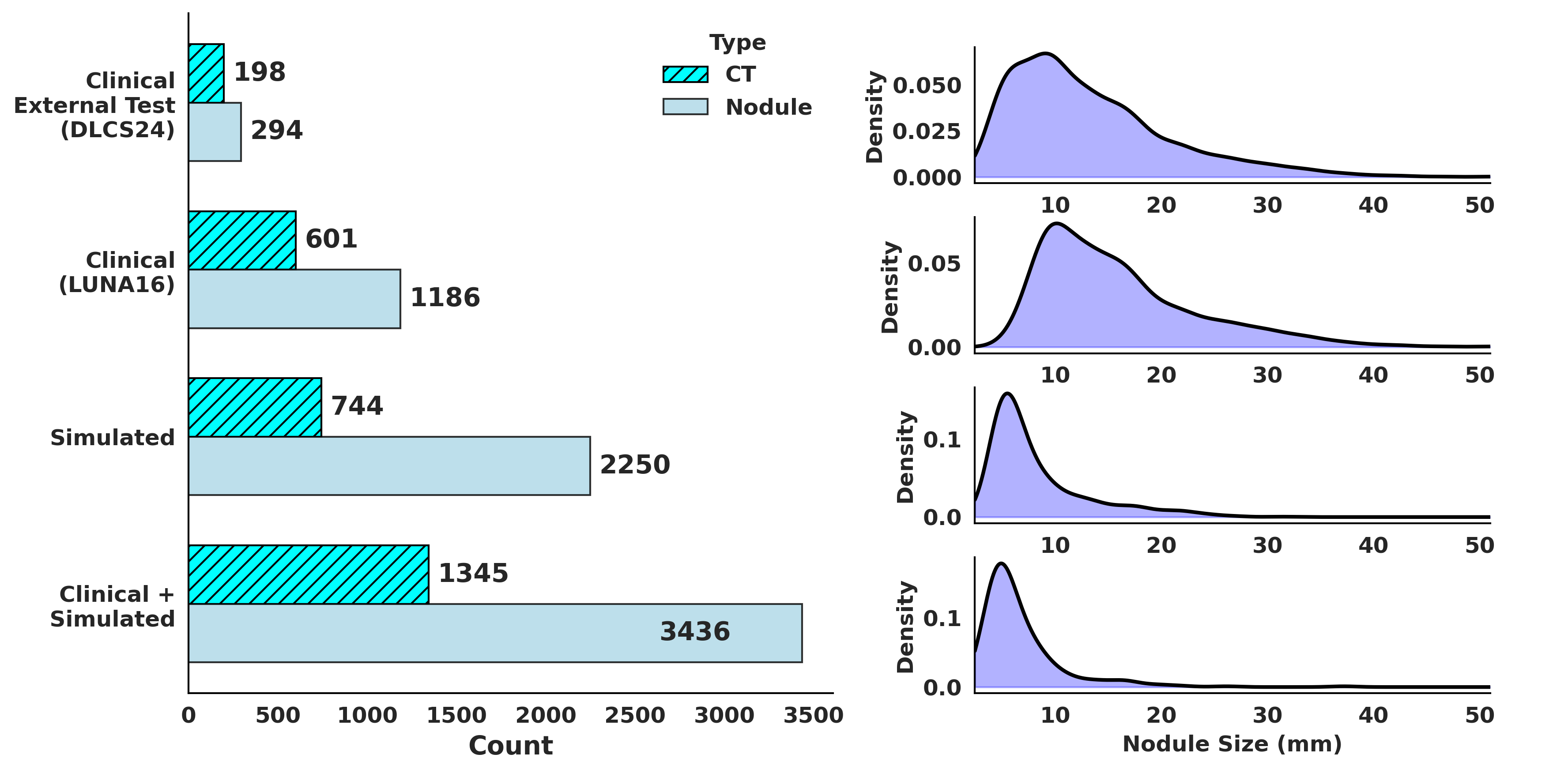}
        \caption{Nodule detection (Task 1) dataset composition and size distribution. (Left) Distribution of CT scans and nodules across clinical (LUNA16) and simulated training datasets, showing the addition of 744 simulated scans to 601 clinical scans. (Right) Nodule size distribution (diameter in mm) comparing clinical and simulated datasets. The simulated dataset provides controlled enrichment across all size ranges with emphasis on small nodules (4--10 mm) that are clinically challenging to detect.}
        \label{fig:detect_dataset}
    \end{minipage}\hfill
    \begin{minipage}[t]{0.38\textwidth}
        \centering
        \includegraphics[width=\textwidth]{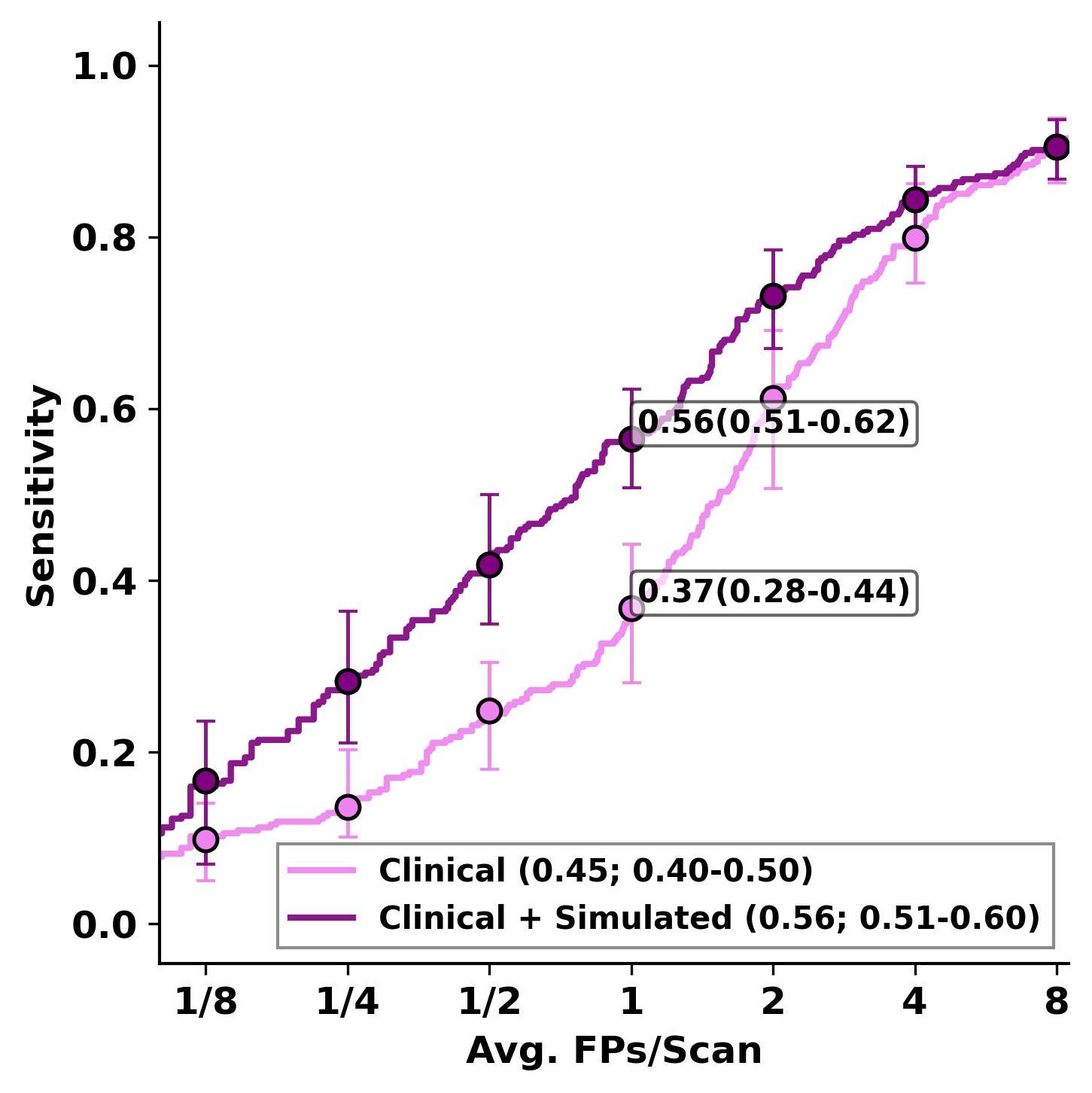}
        \caption{FROC curves for nodule detection on the DLCS24 external test dataset. Clinical-only (light purple/pink) versus clinical + simulated (dark purple) models. Filled circles indicate sensitivity at predefined false-positive rates (1/8--8 FP/scan) with error bars showing 95\% bootstrap confidence intervals.}
        \label{fig:detect_froc}
    \end{minipage}
\end{figure}

\subsection{Task 1: Nodule Detection}
Nodule detection localizes suspicious lesions in 3D CT volumes and outputs bounding boxes. The clinical-only model trained on LUNA16 (601 CTs, 1,186 nodules); the clinical + simulated model added 744 simulated CTs generated under varying scanner and filter settings (Fig.~\ref{fig:detect_dataset}). We used the MONAI detection framework~\cite{cardoso2022monai,tushar2024ai}, a two-stage 3D RetinaNet with region-proposal candidate generation and refined classification. Pre-processing resampled to 0.7 $\times$ 0.7 $\times$ 1.25 mm with intensity normalization; models used 192 $\times$ 192 $\times$ 80 patches via sliding-window inference. Both configurations trained with identical hyperparameters for 200 epochs, selecting the best validation checkpoint. Detection was evaluated on the external DLCS24 test set (198 CTs)~\cite{wang2025duke,tushar2024ai} using free-response receiver operating characteristic (FROC) analysis, reporting sensitivity at 1/8--8 FP/scan and summarizing average sensitivity as the Competition Performance Metric (CPM)~\cite{setio2017validation}. Adding simulated data improved detection at every operating point (Fig.~\ref{fig:detect_froc}). Sensitivity at 1 FP/scan rose from 0.37 (95\% CI 0.28--0.44) to 0.56 (0.51--0.62; p < 0.001), and CPM improved from 0.45 (0.40--0.50) to 0.56 (0.51--0.60). Gains were most pronounced at low false-positive rates, where small-nodule detection is hardest.

\begin{table}[!t]
\centering
\caption{Segmentation Dice performance on MSD test datasets. Dice scores are reported as mean $\pm$ standard deviation and median (interquartile range).}
\label{tab:Seg_dice}
\scriptsize
\setlength{\tabcolsep}{3pt}
\renewcommand{\arraystretch}{1.1}
\begin{tabular}{|l|c|c|c|c|}
\hline
 & \multicolumn{2}{c|}{\textbf{VISTA3D}} 
 & \multicolumn{2}{c|}{\textbf{nnUNet}} \\
\cline{2-5}
\textbf{Training Regime}
& \textbf{Mean $\pm$ SD} 
& \textbf{Median (IQR)}
& \textbf{Mean $\pm$ SD} 
& \textbf{Median (IQR)} \\
\hline

Clinical
& 0.61 $\pm$ 0.34 & 0.73 (0.36--0.87)
& 0.66 $\pm$ 0.33 & 0.79 (0.71--0.86) \\
\hline
Clinical + Simulated
& \textbf{0.64 $\pm$ 0.32} & \textbf{0.76 (0.54--0.86)}
& \textbf{0.69 $\pm$ 0.33} & \textbf{0.85 (0.76--0.88)} \\
\hline
\end{tabular}
\end{table}

\subsection{Task 2: Nodule Segmentation}

Segmentation delineates 3D nodule boundaries for volume measurement and characterization. We evaluated two complementary models: VISTA3D~\cite{he2025vista3d}, a foundation segmentation model fine-tuned from pretrained weights, and nnU-Net(v2)~\cite{isensee2021nnu}, trained from scratch with automatic self-configuration. The clinical-only configuration used MSD Task06 (50 CTs with voxel-wise masks); the clinical + simulated configuration added simulated data with complete 3D annotations. Both models shared preprocessing (resampling, intensity normalization) and training protocol, with best-model selection on validation Dice. Performance was evaluated on the MSD Task06 held-out test set. Figure~\ref{fig:seg_dataset}  shows nodule-size distributions across clinical, simulated, and combined data, highlighting the enrichment of smaller nodules from simulation. Simulated enrichment produced modest but consistent gains across both architectures (Table~\ref{tab:Seg_dice}). VISTA3D mean Dice rose from 0.61 $\pm$ 0.34 to 0.64 $\pm$ 0.32 (median 0.73 $\rightarrow$ 0.76), and nnU-Net from 0.66 $\pm$ 0.33 to 0.69 $\pm$ 0.33 (median 0.79 $\rightarrow$ 0.85). The small test set (13 nodules) limits statistical power, so these results are best read as a favorable trend rather than a definitive gain.

\subsection{Task 3: Malignancy Classification}

\begin{figure}[!t]
   \centering
   \begin{minipage}[t]{0.60\textwidth}
      \centering
      \includegraphics[width=\linewidth]{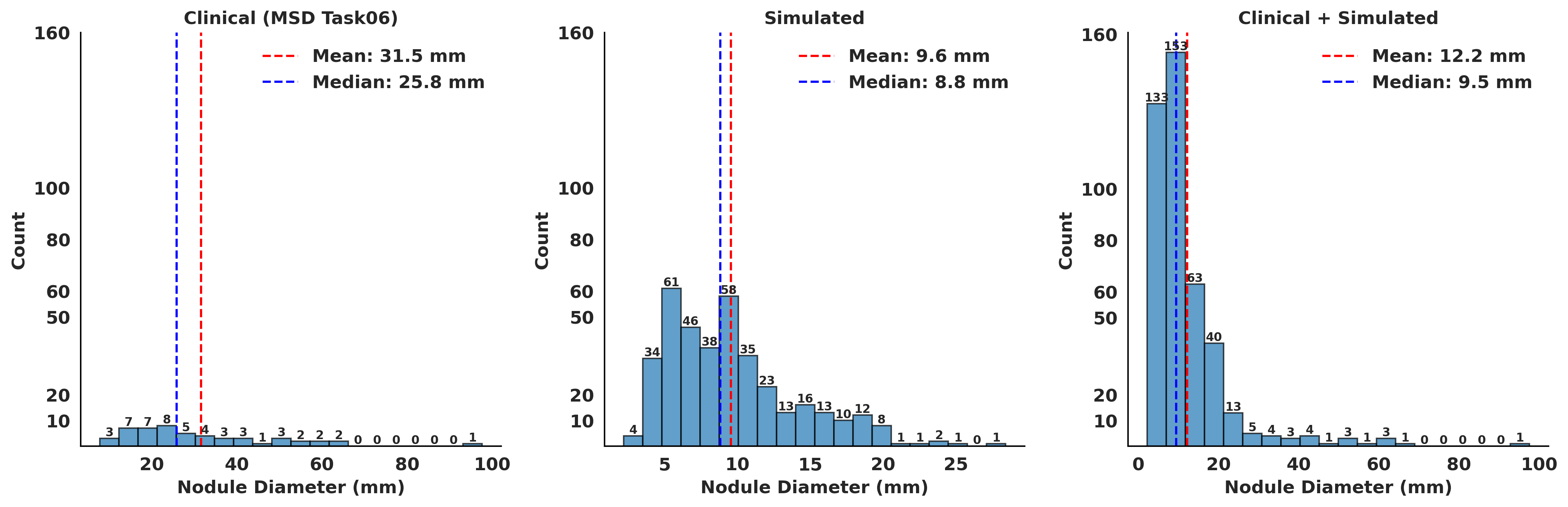}
      \caption{Nodule-size distributions for clinical, simulated, and combined segmentation datasets. Dashed lines indicate mean and median diameters.}
      \label{fig:seg_dataset}
   \end{minipage}\hfill
   \begin{minipage}[t]{0.30\textwidth}
      \centering
      \includegraphics[width=\linewidth]{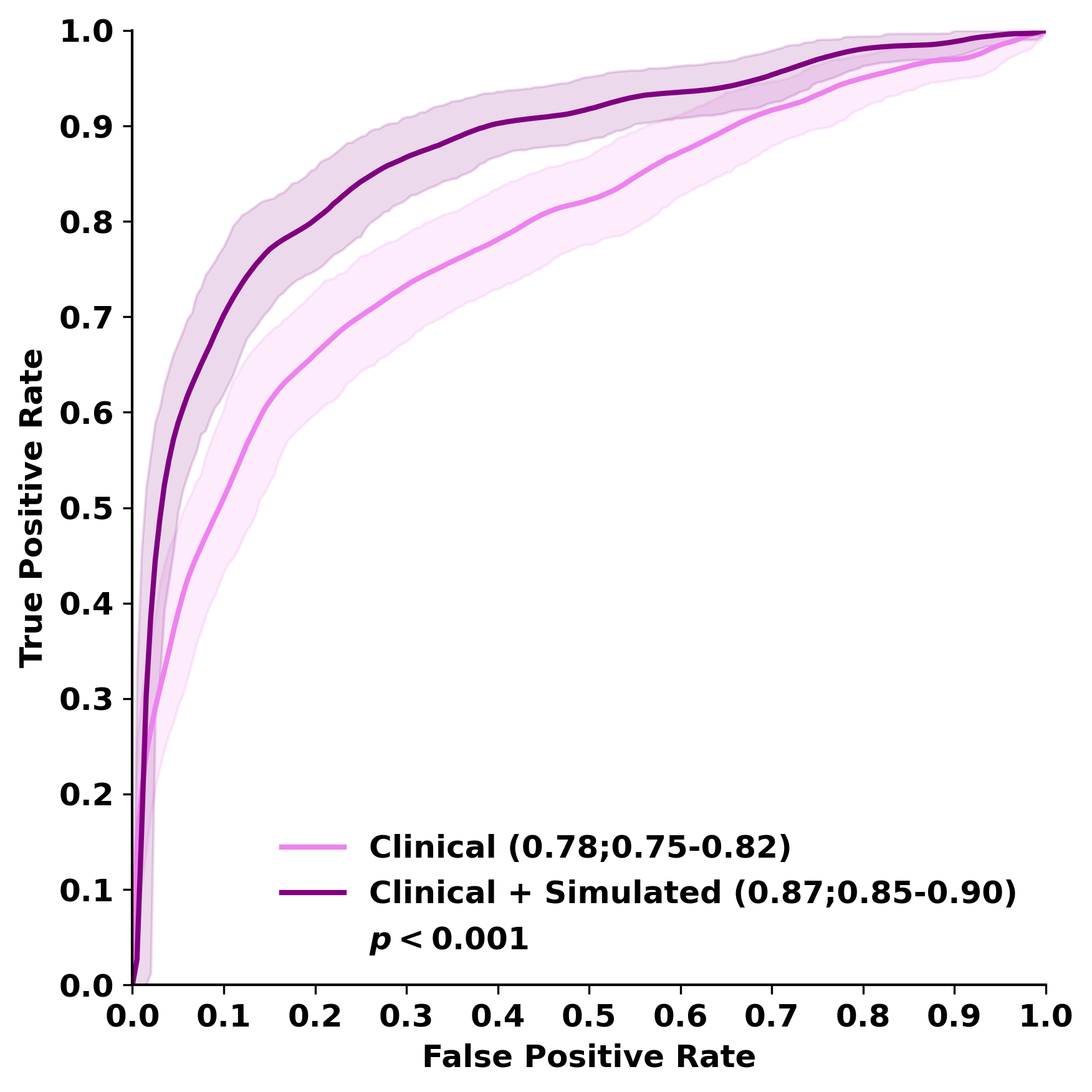}
      \caption{ROC curves for malignancy classification on LUNA16, comparing clinical-only and clinical + simulated training.}
      \label{fig:cls_auc}
   \end{minipage}
\end{figure}

Classification distinguishes malignant from benign nodules on CT. The clinical-only model used DLCS24 (1,618 training nodules: 1,452 benign, 166 malignant; 575 validation)~\cite{tushar2024ai,wang2025duke}; the clinical + simulated model added 366 simulated nodules (approximately 23\% of the clinical training size). Simulated-nodule malignancy labels came from a logistic-regression model adapted from prior work~\cite{tushar2024beyond}, trained on NLST features (age, sex, nodule size, margin, location, type); it achieved AUC 0.72 (95\% CI 0.71--0.74) on both NLST train and test splits, indicating stable calibration for label generation while acknowledging the absence of biological modeling. We fine-tuned Med3D~\cite{chen2019med3d}, a 3D ResNet50 pretrained on medical imaging. Volumes were resampled to 0.7 $\times$ 0.7 $\times$ 1.25 mm, intensity-clipped (-1000 to 500 HU) and standardized, then cropped to 64 $\times$ 64 $\times$ 64 patches centered on nodule centroids. Training ran 200 epochs with best-model selection on validation AUC. Classification was evaluated on external LUNA16 (677 RSL nodules~\cite{pai2024foundation,tushar2024ai}) using AUC with bootstrap 95\% CIs (2,000 resamples) and DeLong testing~\cite{delong1988comparing}. Adding simulated data raised classification AUC from 0.78 (0.75--0.82) to 0.87 (0.85--0.90; p < 0.001) (Fig.~\ref{fig:cls_auc}), the largest relative improvement of the three tasks.

\section{Discussion}
Enriching clinical training data with physics-based, anatomy-informed simulated CT improved AI performance across all three tasks, with substantial gains in detection and classification and a smaller, likely underpowered improvement in segmentation. In every case the clinical + simulated model matched or improved upon the clinical-only baseline on external test sets, supporting virtual imaging trials as a practical route to addressing data scarcity in medical AI development.

Our approach differs fundamentally from data-driven generative synthesis. Where diffusion- or GAN-based methods learn image distributions without explicit anatomical grounding, our framework combines validated computational phantoms~\cite{dahal2025xcat}, procedural lesion models~\cite{tushar2022virtual,mccabe2024synthesizing}, and physics-based imaging simulation~\cite{abadi2018dukesim}, yielding anatomically consistent images with complete, exact ground truth. This control enables systematic coverage of the size and acquisition space---including the small nodules and low-false-positive regime where detection is hardest---that generative models cannot reliably guarantee.

Several limitations merit consideration. The simulated cohort derives from 174 unique digital twins, bounded by the computational cost of full-chest modeling~\cite{tushar2022virtual}. Current nodules are restricted to solid types; extension to semi-solid and ground-glass presentations is underway. The CT simulation replicates NLST-era scanners and should be extended to modern technologies including photon-counting detectors. Statistical malignancy labeling, while demonstrating stable calibration, lacks biological growth modeling; future work will incorporate nutrient-diffusion and angiogenesis models~\cite{tushar2024beyond}. Finally, the segmentation test set was small (13 nodules), limiting statistical power; nevertheless, this dataset has been widely used for evaluation in prior work~\cite{guo2025maisi,he2025vista3d}.

These findings point to several future directions: systematic study of augmentation strategies (mixing ratios, nodule-type balance), extension to multi-disease modeling, incorporation of temporal dynamics for longitudinal prediction, and AI-driven protocol optimization. As digital-twin generation becomes more efficient, virtual imaging trials should play a growing role in developing reliable, generalizable AI systems for medical imaging. All code, models, and protocols are publicly available to support reproducibility and community development.

\section{Conclusion}

Physics-based, anatomy-informed simulated CT data can improve AI performance for lung cancer diagnosis. Adding simulated data to clinical training yielded clear gains in nodule detection and malignancy classification and a modest improvement in segmentation, evaluated on external test sets. Virtual imaging trials thus offer a viable response to data scarcity, providing anatomically realistic augmentation with complete ground truth. As digital-twin generation matures, this approach points toward more reliable AI systems for lung cancer screening.

\begin{credits}
\subsubsection{\ackname}
This work was funded in part by the National Institutes of Health (P41-EB028744, R01EB001838, R01HL155293). We also thank MONAI and NVIDIA for providing open-access code resources that supported this research.

\subsubsection{\discintname}
The authors have no competing interests relevant to the content of this article.
\end{credits}

\end{document}